\documentclass[conference]{IEEEtran}
\IEEEoverridecommandlockouts
\usepackage{cite}
\usepackage{amsmath,amssymb,amsfonts}
\usepackage{algorithmic}
\usepackage{graphicx}
\usepackage{multirow}
\usepackage{multicol}
\usepackage{booktabs}
\usepackage{makecell}
\usepackage{subcaption}
\usepackage{comment}
\usepackage[colorlinks=true, allcolors=blue]{hyperref}

\usepackage{textcomp}
\usepackage{xcolor}
\def\BibTeX{{\rm B\kern-.05em{\sc i\kern-.025em b}\kern-.08em
    T\kern-.1667em\lower.7ex\hbox{E}\kern-.125emX}}
\begin{document}

\title{Efficient Curation of Invertebrate Image Datasets Using Feature Embeddings and Automatic Size Comparison\\
\thanks{This work was funded by Research Council of Finland project 333497.}
}
\author{\IEEEauthorblockN{Mikko Impiö}
\IEEEauthorblockA{\textit{Quality of Information} \\
\textit{Finnish Environment Institute}\\
Helsinki, Finland \\
mikko.impio@syke.fi}
\and
\IEEEauthorblockN{Philipp M. Rehsen}
\IEEEauthorblockA{\textit{Aquatic Ecosystem Research, Faculty of Biology} \\
\textit{Centre for Water and Environmental Research (ZWU)}\\
\textit{University of Duisburg-Essen}\\
Essen, Germany \\
philipp.rehsen@uni-due.de}
\and
\IEEEauthorblockN{Jenni Raitoharju}
\IEEEauthorblockA{\textit{Faculty of Information Technology} \\
\textit{University of Jyväskylä}\\
Jyväskylä, Finland \\
jenni.k.raitoharju@jyu.fi}
}

\maketitle

\begin{abstract}
The amount of image datasets collected for environmental monitoring purposes has increased in the past years as computer vision assisted methods have gained interest.
Computer vision applications rely on high-quality datasets, making data curation important. However, data curation is often done ad-hoc and the methods used are rarely published.
We present a method for curating large-scale image datasets of invertebrates that contain multiple images of the same taxa and/or specimens and have relatively uniform background in the images.
Our approach is based on extracting feature embeddings with pretrained deep neural networks, and using these embeddings to find visually most distinct images by comparing their embeddings to the group prototype embedding.
Also, we show that a simple area-based size comparison approach is able to find a lot of common erroneous images, such as images containing detached body parts and misclassified samples.
In addition to the method, we propose novel metrics for evaluating human-in-the-loop outlier detection methods.
The implementations of the proposed curation methods, as well as a benchmark dataset containing annotated erroneous images, are publicly available in \texttt{\url{https://github.com/mikkoim/taxonomist-studio}}.
\end{abstract}


\section{Introduction}
\label{sec:introduction}

The need for global biodiversity monitoring efforts is widely recognized, as the state of natural environments continues deteriorating worldwide and major fractions of biodiversity remain unknown \cite{wagner2020Insect, ipbes2019Summary, stork2018How}.
Insects and other arthropods are groups of special interest in these monitoring efforts.
Insects are the most diverse animal group and of great importance in food webs and as providers of ecosystem services \cite{noriega2018Research}.
Monitoring these groups with traditional methods is laborious, and approaches using new technologies like computer vision have been proposed in the recent years \cite{hoye2021Deep}.
Computer vision can address challenges such as species classification, enumeration, and biomass estimation.

Computer vision methods usually rely on deep learning, which in turn needs a lot of training data.
Several approaches for digitizing arthropod samples for computer vision purposes have emerged in the past few years, making collection of larger datasets easier \cite{schneider2022Bulk, deschaetzen2023Riverine, arje2020Automatic, wuhrl2022DiversityScanner, simovic2024Automated}.
The amount of image data is thus growing, but often raw datasets collected with imaging devices contain erroneous images.
These images might be debris, air bubbles, or anything else that does not match the taxonomic class label given to the image.
Removing these images through curation is important in order to produce high-quality training datasets, which usually lead to better classification performance \cite{huh2016What, vanooijen2019Quality}.

Dataset curation remains a laborious process.
In practice, most curation solutions are developed as custom one-off approaches for specific datasets and problems.
Often curation work is also done manually, which can be very time-consuming for large datasets.
Some curation studies exist in the medical imaging domain \cite{vanooijen2019Quality} and for content-based curation of datasets \cite{vannoord2023Prototypebased}, but most curation solutions are not published.

\begin{figure}[bt]
    \centering
    \begin{subfigure}[t]{0.3\columnwidth}
        \centering
        \includegraphics[height=1in]{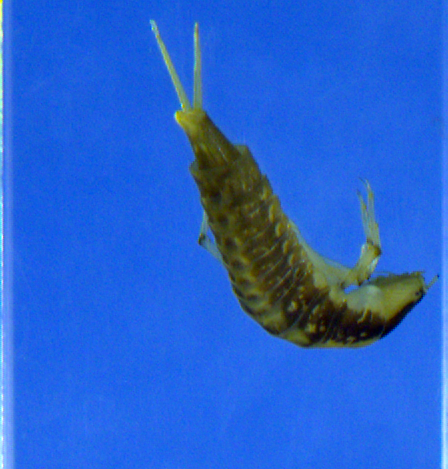}
        \caption{\textit{Ilybius}}
        \label{sfig:seeds}
    \end{subfigure}
    \begin{subfigure}[t]{0.3\columnwidth}
        \centering
        \includegraphics[height=1in]{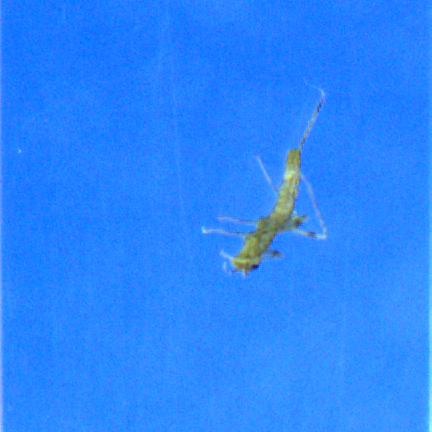}
        \caption{\textit{Ischnura elegans}}
        \label{sfig:ionosphere}
    \end{subfigure}
    \begin{subfigure}[t]{0.3\columnwidth}
        \centering
        \includegraphics[height=1in]{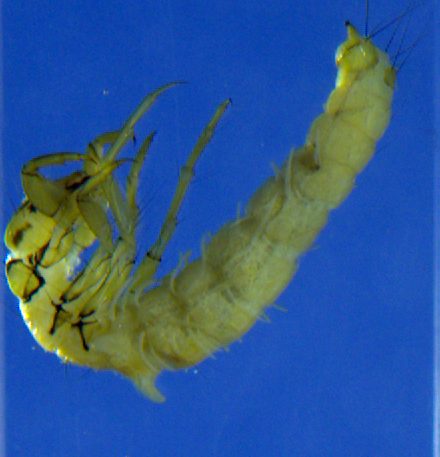}
        \caption{\textit{Phryganea}}
        \label{sfig:average}
    \end{subfigure}
    \caption{Examples of successful BIODISCOVER images}
    \label{fig:goodexamples}
\end{figure}

\begin{figure*}[h!]
\centering
    \centering
    \includegraphics[width=\linewidth]{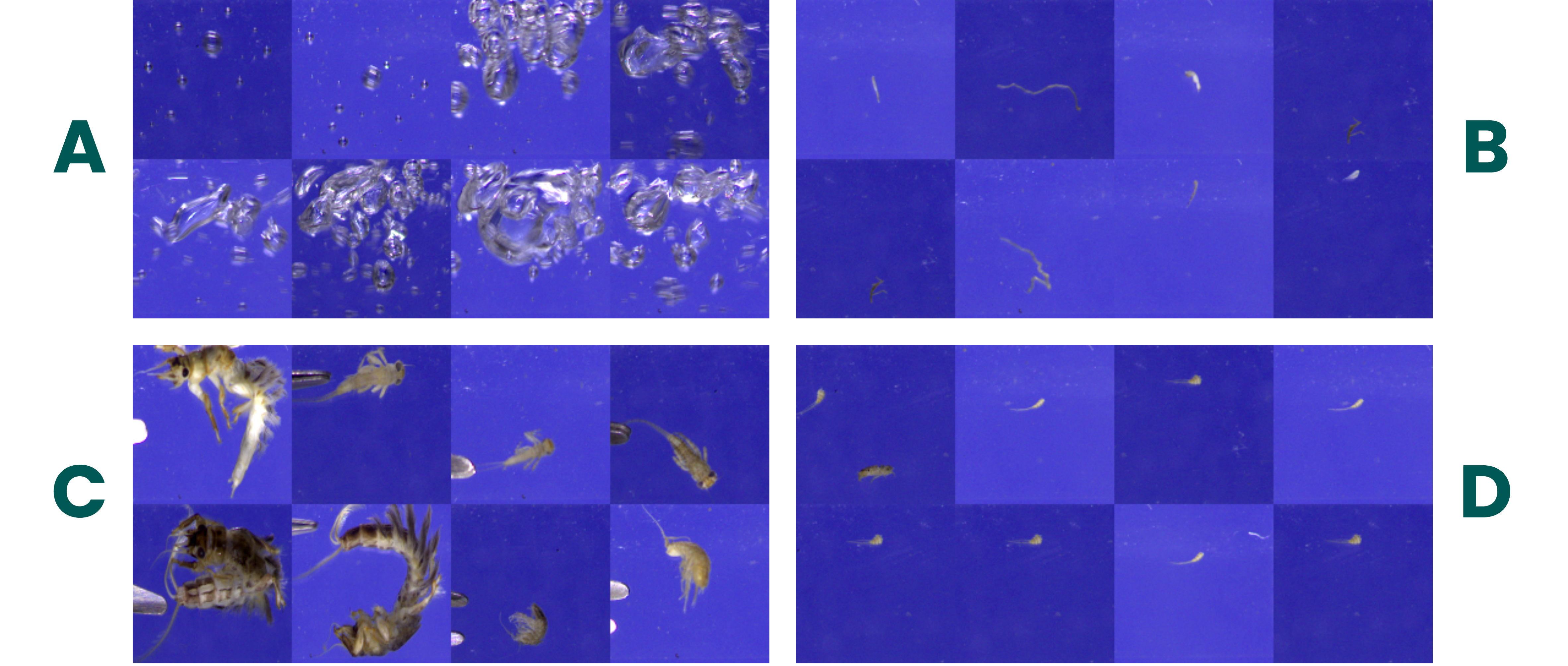}
    \caption{Examples of erroneous content in images captured by a BIODISCOVER imaging device: A: Bubbles, B: Detached body parts, C: Forceps, D: Misclassifications}
    \label{fig:example_images}
\end{figure*}

In this paper, we propose methodology for curating large-scale arthropod datasets imaged with a device, such as BIODISCOVER \cite{arje2020Automatic}, that (semi-)automatically takes multiple images of presorted samples.
A couple of examples of successfully imaged arthropods using the BIODISCOVER imaging device are shown in Fig.~\ref{fig:goodexamples}.
The curation needs may arise from, e.g., debris or misclassified samples among the samples taxonomically presorted  by a human expert, detached body parts, or air bubbles in the ethanol used for imaging the wet samples.
Examples of different types of erroneous content are shown in Fig.~\ref{fig:example_images}. 

Our proposed curation approach is especially useful for initial detection of erroneous content, where a human 
expert makes the final decision on the removal of the detected candidates.
We show that good results can be achieved using deep networks pretrained on unrelated generic datasets and, thus, the approach does not require any training.
The approach is suitable for any dataset that contains multiple images of the same taxon and/or specimens and have relatively uniform background in the images.

In addition to the proposed dataset curation approach, our contributions include publishing a novel real-life dataset suitable for evaluating data curation approaches and proposing three novel metrics suitable for the evaluation task.

The rest of the paper is organized as follows. In Section~\ref{sec:methods}, we present our approach feature embedding and size-based dataset curation. In Section~\ref{sec:noveldata}, we introduce the novel benchmark dataset. The proposed novel metrics are introduced in Section~\ref{sec:experiments} along with overall experimental setup and experimental results that show the effectiveness of the proposed dataset curation approach via experiments on two real-life arthropod datasets.
Finally, in Section ~\ref{sec:discussion}, we conclude the paper.

\section{Proposed Dataset Curation Approach}
\label{sec:methods}

We use two different methods for dataset curation: the first is based on image content comparison using feature embeddings and the second is based on size comparison.
Both of the methods are designed for datasets that contain multiple images representing a single individual and/or taxon.

\subsection{Dataset Curation via Feature Embeddings}
Image content can be compared by calculating feature embeddings for the images in the dataset.
The embeddings are dense, real valued vectors representing the original images and their contents. We propose extracting the feature embeddings from a pretrained deep neural network that consists of a feature encoder $f$ and a classification head $g$, so that their composition maps image inputs $\mathbf{x}_i$ from a dataset $\mathcal{X} = \{\mathbf{x}_i\}_{i=1}^N$ to a target class $y_i\in \mathcal{Y}$, i.e., $g\circ f: \mathcal{X} \rightarrow \mathcal{Y}$.
Both $f$ and $g$ can be deep neural networks, however $g$ usually consists of a single fully-connected feed-forward layer.
The intermediate output $\mathbf{z}_i = f(\mathbf{x}_i), \mathbf{z}_i \in \mathbb{R}^M$ is an $M$-dimensional vector usually called a \textit{feature vector}, or an \textit{embedding}. The similarity of two feature embeddings $\mathbf{z}_i$ and $\mathbf{z}_j$ can be compared by calculating their cosine distance from each other.
Cosine distance for vectors $\mathbf{z}_i$ and $\mathbf{z}_j$ is
\begin{equation}
    d_{\cos}(\mathbf{z}_i,\mathbf{z}_j) = 1- \frac{\mathbf{z}_i\cdot \mathbf{z}_j}{||\mathbf{z}_i||_2||\mathbf{z}_j||_2},
\end{equation}
meaning that orthogonal vectors have a distance of 1. 

In our experiments, we use feature embeddings extracted from a simple MobileNet-v3 \cite{howard2019Searching} model pretrained on ImageNet-1k dataset \cite{russakovsky2015ImageNet}. However, the data curation approach can be used with any deep neural network model pretrained on any dataset. Our experimental results demonstrate that a network pretrained using a generic unrelated dataset can produce meaningful feature embeddings for our dataset curation approach, i.e., no training on the dataset to be curated is required.

After extracting the feature embeddings for each dataset image, we choose a grouping (for example based on a taxon or a specimen) for which a mean feature embedding is calculated, and each image (feature embedding) in the same group is compared to.
This is similar to calculating a "prototype" vector for a group, as formulated in \cite{snell2017Prototypical}.
We assume that this mean feature embedding encodes some properties of the group and meaningful distances to it can be calculated, although we do not explicitly use a metric learning approach as in \cite{snell2017Prototypical}.

For a group $c$, we calculate the mean feature embedding by taking all images that belong to this group $\mathcal{X}^c  = \{\mathbf{x}_i^c\}_{i=1}^{N^c}$, and calculating their embeddings, so that we have a set of feature embeddings $\mathcal{Z}^c = \{\mathbf{z}_i^c\}_{i=1}^{N^c}$, where $N^c$ is the amount of images in group $c$.
The formula for computing the a mean feature embedding for group $c$, $\mathbf{m}^c$, is
\begin{equation}
\label{eq:prototype_vector}
    \mathbf{m}^c = \frac{1}{N^c}\sum_{i=1}^{N^c} \mathbf{z}_i^c.
\end{equation}

Now, the distances of all feature embeddings in $\mathcal{Z}^c$ to their group mean vector, $\mathbf{m}^c$, can be separately calculated:
\begin{equation}
    d_i^c = d_{\cos}(\mathbf{z}_i^c, \mathbf{m}^c).
\end{equation}
If we interpret the mean feature embedding to be a "prototype" representing the entire group, images whose feature embeddings are far away from it are likely to be outliers having erroneous content.
Therefore, by ordering the images by their distance in descending order, we can find the most dissimilar images to the mean embedding.
These images can be then manually inspected and removed from the dataset if they are indeed detected to have unwanted content.
The inspection can be performed either separately for each group or by pooling samples from all groups together.
Our approach assumes that the majority of images in each group are of good-quality.
If erroneous images are in majority, the good-quality images will be ranked first.
This might be true for smaller groupings, for example on sample level, but is very unlikely with larger groupings.

It is easy to order images inside their respective groups, as all distances are calculated to the same mean embedding $\mathbf{m}^c$.
However, if one wants to inspect the full dataset instead of groups, ordering can be done for the full dataset.
When pooling across groups, some groups have smaller variance among their elements, and possible outliers might be still relatively close to the group mean.
On the other hand, groups with a larger variance have large distances to the group mean feature embedding also from high-quality images. Thus, if the distances are directly ranked across all groups, high-quality images in high-variance groups can be ranked before outliers in low-variance groups, making dataset curation difficult. We can correct this by normalizing each distance in the group by the mean distance in the group:
\begin{equation}
    d_i^c = \frac{d_{\cos}(\mathbf{z}_i^c, \mathbf{m}^c)}{\bar{d^c}},
\end{equation}
where
\begin{equation}
    \bar{d^c} = \frac{1}{{N^c}^2}\sum_{i=1}^{N^c}\sum_{j=1}^{N^c} d_{\cos}(\mathbf{z}_i^c,\mathbf{z}_j^c).
\end{equation}

\subsection{Dataset Curation via Size Comparison}

In addition to data curation via feature embedding comparison, we detect outliers by comparing the image areas to the mean area of a comparison group. Intuitively, this approach can be suitable for detecting, e.g., pictures of detached body parts that have been imaged due to sensitive motion detection in the imaging system.

Each image $\mathbf{x}_i$ has a corresponding area $a_i$, which is the area (in pixels) the specimen takes up in the image.
The specimen area is calculated by comparing each frame to a calibration image containing only the empty background.
The difference between each frame and the calibration image is calculated, and pixels where the difference is higher than a threshold value are considered as specimen pixels.
After applying morphological operations and filtering the largest blob is chosen as the final object and its area as the specimen area.

The mean area in group $c$ can be computed as 
\begin{equation}
    \bar{a}^c = \frac{1}{N^c}\sum_{i=1}^{N^c} a_i^c,
\end{equation}
where $a_i^c$ is the area of image $\mathbf{x}_i^c$.

We can then calculate a percentual difference for each image $\mathbf{x}_i^c$ by
\begin{equation}
   a_{i\Delta}^c = \frac{\left| a_i^c - \bar{a}^c \right|}{\bar{a}^c}.
   \label{eq:areadifference}
\end{equation}
We rank all images based on $a_{i\Delta }^c$ so that largest values are the most probable outlier candidates.

\section{Novel Benchmark Dataset for Curation Approaches}
\label{sec:noveldata}

In order to test our dataset curation approach, we collected and manually annotated a dataset, available at \texttt{\url{https://github.com/mikkoim/taxonomist-studio}}. The dataset was collected with an updated version of BIODISCOVER device \cite{arje2020Automatic}, which captures image sequences of a specimen as it falls through an ethanol-filled cuvette. Images are captured from two perpendicular angles. Examples of high-quality images with slight brightness and contrast adjustment for illustration purposes are shown in Fig.~\ref{fig:goodexamples}.

\begin{figure}[h!]
  \includegraphics[width=0.9\linewidth]{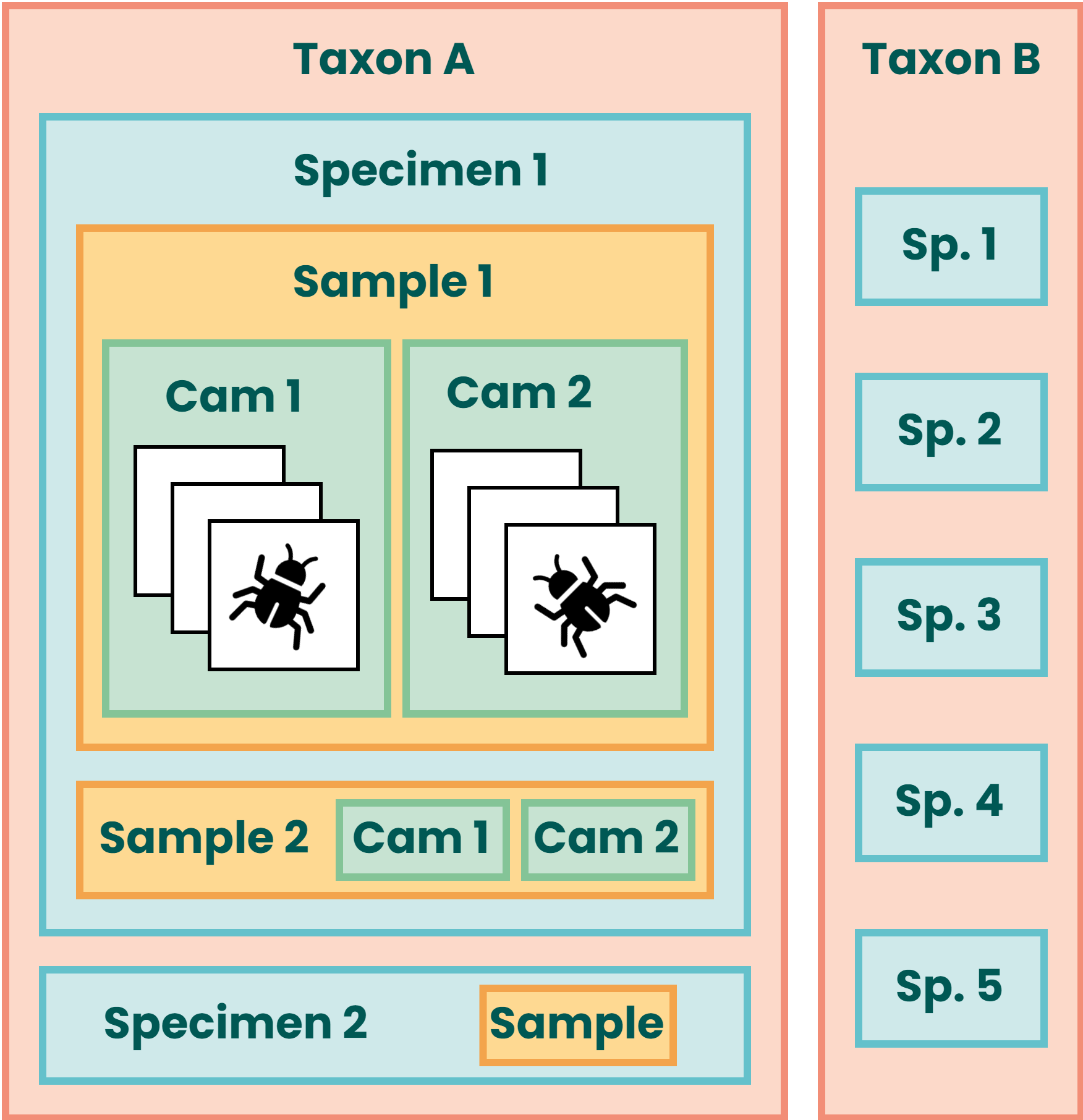}
  \caption{Different possible groupings for our datasets. Each individual \textit{Specimen} (Sp) belongs to a \textit{Taxon} (T). When a specimen is imaged, a single imaging run is called a \textit{Sample} (Sa), which consists of two image sequences from two camera angles. We refer to a sequence from one of the cameras as the \textit{Cam} (C) sequence.}
  \label{fig:groupings}
\end{figure}

Each individual \textit{specimen} might be imaged several times -- we call one of these imaging runs a \textit{sample}.
Each sample produces an image sequence from two cameras, referred to as the \textit{cam} sequences.
Overview of the hierarchy of different image groupings is illustrated in Fig.~\ref{fig:groupings}.
These groupings are also the different groupings we use as $c$ in Eqs.~\eqref{eq:prototype_vector}-\eqref{eq:areadifference}.

The dataset has a total of 90380 images from 24 categories with 406 true outlier image annotations.
These images are from 3107 samples of 1518 specimens. The number of images per specimen was capped to the maximum of 50 images per camera angle representing an imaging run. The outliers were naturally produced during a regular imaging campaign and, thus, represent the realistic scenario in terms of frequency and types of different outliers.
These outliers were classified to four different classes: Bubbles (N=22), Detached parts (N=324), Forceps (N=31), and Misclassifications (N=29).
Examples of these images can be seen in \ref{fig:example_images}.

\begin{table*}[h!]
\caption{Dataset curation results for the two proposed approaches over two datasets
\label{tab:results}}
\centering
\begin{tabular}{llrrrrr}
\toprule
 Outlier type & Method & AUROC~$\uparrow$ & AP~$\uparrow$ & TPR@Head~$\uparrow$ & Rec@5\%p~$\uparrow$ & p@95\%Rec.~$\downarrow$ \\
\midrule
\multicolumn{7}{c}{Benchmark dataset (N=90 380)}\\
\midrule
\multirow[c]{2}{*}{\shortstack[l]{All outliers\\\small{(N=406)}}} & Embedding & 89.9 & 14.4 & 17.0 & 62.6 & 54.1 \\
 & Size & 95.3 & 44.0 & 55.9 & 91.6 & 39.4 \\
\cmidrule(lr){1-7}
\multirow[c]{2}{*}{\shortstack[l]{Bubbles\\\small{(N=22)}}}  & Embedding & 100.0 & 81.0 & 77.3 & 100.0 & 0.2 \\
 & Size & 97.9 & 29.2 & 27.3 & 95.5 & 5.0 \\
  \cmidrule(lr){2-7}
\multirow[c]{2}{*}{\shortstack[l]{Detached parts\\\small{(N=324)}}} & Embedding & 94.3 & 6.6 & 11.4 & 66.0 & 24.4 \\
 & Size & 99.8 & 43.3 & 49.1 & 100.0 & 0.6 \\
  \cmidrule(lr){2-7}
\multirow[c]{2}{*}{\shortstack[l]{Forceps\\\small{(N=31)}}}  & Embedding & 91.0 & 1.3 & 3.2 & 58.1 & 41.3 \\
 & Size & 47.4 & 0.0 & 0.0 & 6.5 & 98.7 \\
  \cmidrule(lr){2-7}
\multirow[c]{2}{*}{\shortstack[l]{Misclassification\\\small{(N=29)}}} & Embedding & 31.9 & 0.0 & 0.0 & 0.0 & 87.6 \\
 & Size & 93.1 & 1.8 & 0.0 & 86.2 & 46.6 \\
\midrule
\multicolumn{7}{c}{Additional data (N=163 519)}\\
\midrule
\multirow[c]{2}{*}{\shortstack[l]{All outliers\\\small{(N=9425)}}} & Embedding & 82.7 & 16.0 & 6.3 & 5.5 & 48.8 \\
& Size & 94.4 & 56.1 & 77.8 & 65.0 & 37.4 \\
\end{tabular}
\end{table*}

\section{Experiments}
\label{sec:experiments}

\subsection{Experimental Setup}

We conducted our experiments on the benchmark dataset introduced in Section~\ref{sec:noveldata}. We also repeated the experiments on a second larger dataset, which cannot be published yet due to ongoing research efforts. The second dataset was also collected using the BIODISCOVER device. It does not have outliers divided to separate classes as the public data. The second dataset has a total of 163 519 images, with 9425 known outliers.
These images are from 2677 samples of 743 specimens, without a cap in the number of images a specimen might have.

We calculated the embeddings using a simple MobileNet-v3 \cite{howard2019Searching} model (\texttt{mobilenetv3\_small\_075.lamb\_in1k}) from the PyTorch Image Models library \cite{rw2019timm}, pretrained with the ImageNet-1k dataset \cite{russakovsky2015ImageNet}.
The area calculations were done using the BIODISCOVER imaging software.
Our dataset curation approach was implemented with Python and is available as part of the software package in \texttt{\url{https://github.com/mikkoim/taxonomist-studio}}.

\subsection{Evaluation Metrics}

We used commonly used outlier detection metrics, such as the area under the receiver operating characters curve (AUROC) and average precision (AP) to evaluate our approach. 
In addition, we define three additional metrics useful for evaluating a dataset curation approach in a setup where it can be assumed that a human annotator will go through the outlier candidates. We are interested in how much human effort is needed to remove all/most outliers from an dataset reliably.

The first additional metric we define is the true positive rate at the "head" of the dataset (TPR@Head), or after $N$ samples, where $N$ is the known number of outliers in the dataset.
With a perfect detector, all outliers would be ranked in the head, making the value of this metric $1.0$.
The second additional metric can help the user to evaluate
how large the expected outlier recall would be after inspecting 5\% of the dataset.
We call this the Rec@5\%p metric. With this metric higher values are better. The third additional metric provides information on the percentage $p$ of the dataset one has to go through in order to achieve a outlier recall rate of 95\%.
We call this the p\%@95Rec metric. Here, a lower value means better performance.

\begin{figure*}
  \centering
  \includegraphics[width=0.7\linewidth]{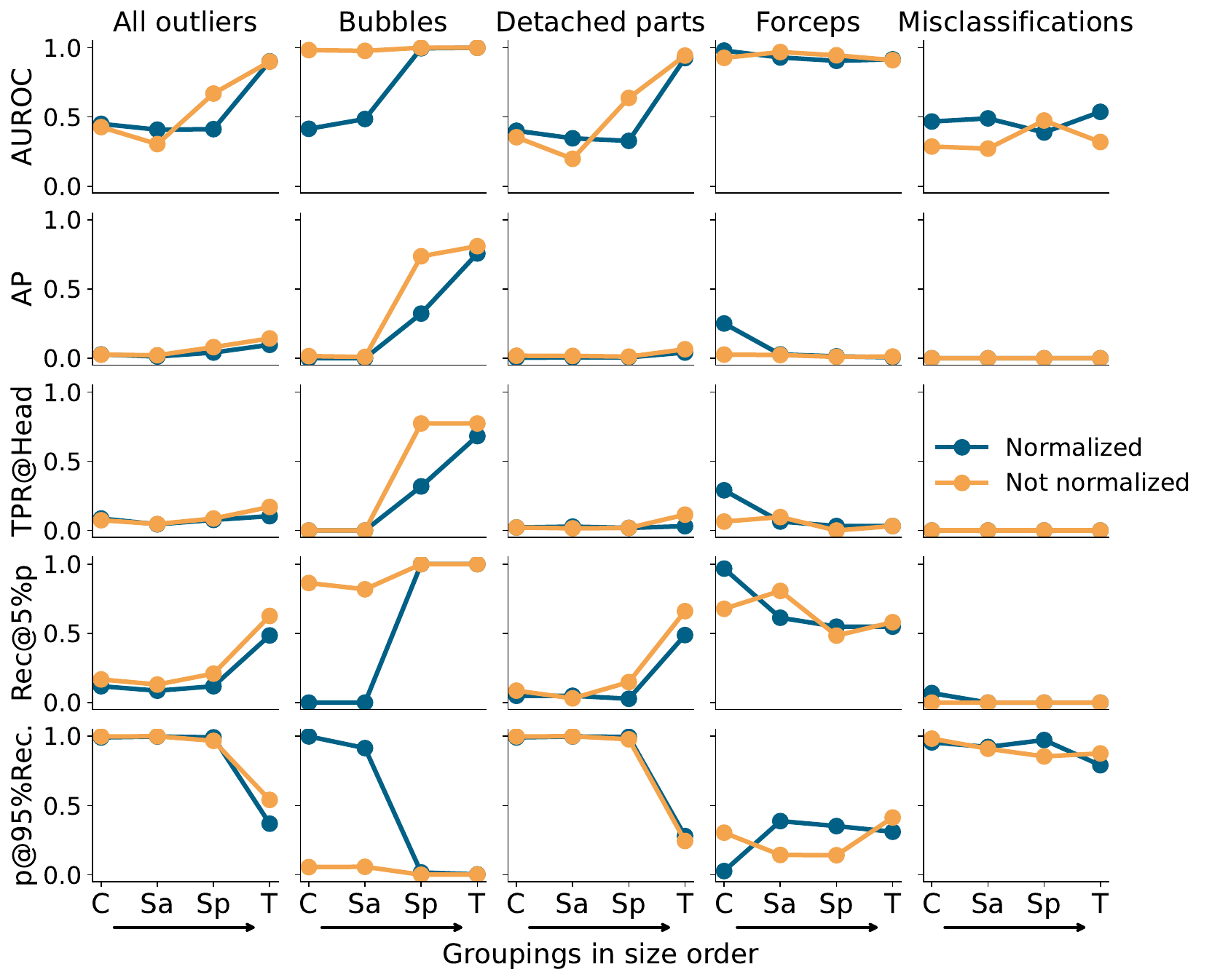}
  \caption{Comparison of the embedding method performance measured with several metrics and for different outlier types. The performance depends on the grouping chosen, seen in the x-axis in increasing group size. C=Cam, Sa=Sample, Sp=Specimen, T=Taxon. Fig.~\ref{fig:groupings} explains the groupings. Overall best results are achieved with the taxon-level grouping.}
  \label{fig:outlier_detection_results}
\end{figure*}

\subsection{Experimental Results}

Table~\ref{tab:results} shows an overview of the results for the two different curation approaches for both datasets, measured with the five different metrics introduced above.
Overall performance of both embedding and size-based approach are similar.
Largest differences can be seen in forcep detection, where the embedding method excels, and in misclassification detection, where the size comparison method is better.
The embedding method is especially good in detecting bubbles from the dataset - after going through 5\% of the dataset, all of the bubble images were found.
Similarly, all detached parts were found with the size comparison method.

The performance of an embedding method depends largely on the group chosen, as it affects the prototype embedding and possible normalization.
Fig.~\ref{fig:outlier_detection_results} shows the performance of both embedding methods on different outlier types and groupings.
The groupings follow the ones seen in Fig.~\ref{fig:groupings}.
For most situations, choosing the comparison group as taxon is the best approach.
Also, normalization does not usually produce better results. It was useful only for finding forceps or for groupings finer than taxon. Here, it should be noted that the taxon-level grouping can be only used for annotated datasets, where as the other groupings can be also when analyzing unknown samples, which would be the case in automated biomonitoring setup.

Our approach uses the cosine distance as the distance function for feature embeddings. We tested also Euclidean distance. The performance was similar to the performance of cosine distance, but cosine distance performed slightly better on most metrics. Thus, cosine distance seems to be suitable distance metric choice for the proposed approach.

\section{Conclusions}
\label{sec:discussion}

In this paper, we presented an easy-to-use approach for curating invertebrate datasets using feature embeddings and size comparisons.
As our results show, the methods can be useful for removing common erroneous content, such as bubbles and detached parts, from image datasets.
The two methods are complementary to each other - the embedding method performs well in finding image content-based outliers, such as forceps and bubbles, whereas size-based comparison is good especially for detached body parts and misclassified samples.

Images of detached body parts are especially common in invertebrate image datasets like ours. The collected specimens can quite easily fall apart during imaging and a detached body part, such as a leg, typically falls through the imaging cuvette significantly slower than the actual specimen.
Because the imaging device automatically captures images of all moving objects in the imaging area, after the actual specimen is imaged and has vanished from the camera view, the software starts imaging the slowly falling detached body part.
This produces a lot of extra images in the same sequence as the images from the actual specimen.
Our results show that these outliers are easy to remove - just going through 5\% of the dataset using the proposed curation approach finds most or all detached part images.

Our dataset curation approach is  suited for automatically collected datasets that contain multiple images of the same taxa and/or specimens and have relatively uniform background in the images. Besides the BIODISCOVER used in our experiments, such datasets could be produced, e.g., by camera traps. The proposed curation approach is not expected to be successful with image datasets that have large variance in the backgrounds, e.g., images taken with camera phones.

In addition to the proposed dataset curation approach, we published a novel real-life dataset suitable for evaluating data curation approaches and proposed three novel metrics suitable for the evaluation task. These will support also others who may want to develop more advanced methods for automatic dataset curation.

In the future, we will use the proposed approach in curating large-scale arthropod datasets. We plan to improve the software during future data collection efforts to suite the needs of computer vision based entomology research.
 
\bibliographystyle{ieeetr}
\bibliography{bibliography}
\end{document}